\pdfoutput=1

\documentclass[11pt]{article}

\usepackage[]{ACL2023}

\usepackage{times}
\usepackage{latexsym}
\usepackage{enumitem}
\usepackage{fvextra}
\usepackage{colortbl}
\usepackage{xcolor}
\usepackage[T1]{fontenc}
\usepackage{graphicx}
\usepackage{array}
\usepackage{microtype}
\usepackage{spverbatim}
\usepackage{inconsolata}
\usepackage{xspace}
\usepackage{booktabs}
\usepackage{amsmath}
\usepackage[utf8]{inputenc}
\newcommand{\titlename}{\textsc{Halu-J}\xspace}

\newenvironment{itemize*}%
 {\leftmargini=10pt\begin{itemize}%
  \setlength{\itemsep}{0pt}%
  \setlength{\parskip}{0pt}%
  }%
 {\end{itemize}}
\newenvironment{enumerate*}%
 {\begin{enumerate}%
  \setlength{\itemsep}{0pt}%
  \setlength{\parskip}{0pt}}%
 {\end{enumerate}}

\title{\titlename: Critique-Based Hallucination Judge}

\author{\textbf{Binjie Wang}\textsuperscript{\rm{2,5}}\space\space \space\space \space\space\space
\textbf{Steffi Chern}\textsuperscript{\rm{4,5}}\space\space \space\space \space\space\space
\textbf{Ethan Chern}\textsuperscript{\rm{1,5}}\space\space \space\space \space\space\space
\textbf{Pengfei Liu}\textsuperscript{\rm{1,3,5}}\thanks{~~Corresponding author}\\
    \textsuperscript{\rm 1}Shanghai Jiao Tong University\space\space
    \textsuperscript{\rm 2}Fudan University \\
    \textsuperscript{\rm 3}Shanghai Artificial Intelligence Laboratory \space\space
    \textsuperscript{\rm 4}Carnegie Mellon University \\
    \textsuperscript{\rm 5}Generative AI Research Lab (GAIR)\\
}

\begin{document}
\maketitle
\begin{abstract}


Large language models (LLMs) frequently generate non-factual content, known as hallucinations. Existing retrieval-augmented-based hallucination detection approaches typically address this by framing it as a classification task, evaluating hallucinations based on their consistency with retrieved evidence. However, this approach usually lacks detailed explanations for these evaluations and does not assess the reliability of these explanations. Furthermore, deficiencies in retrieval systems can lead to irrelevant or partially relevant evidence retrieval, impairing the detection process. Moreover, while real-world hallucination detection requires analyzing multiple pieces of evidence, current systems usually treat all evidence uniformly without considering its relevance to the content.
To address these challenges, we introduce \titlename, a critique-based hallucination judge with 7 billion parameters. \titlename enhances hallucination detection by selecting pertinent evidence and providing detailed critiques. Our experiments indicate that \titlename outperforms GPT-4o in multiple-evidence hallucination detection and matches its capability in critique generation and evidence selection. We also introduce \texttt{ME-FEVER}, a new dataset designed for multiple-evidence hallucination detection. Our code and dataset can be found in \url{https://github.com/GAIR-NLP/factool}.


\end{abstract}

\section{Introduction}

\begin{figure}[ht]
  \centering
  \includegraphics[width=\linewidth, height=6cm]{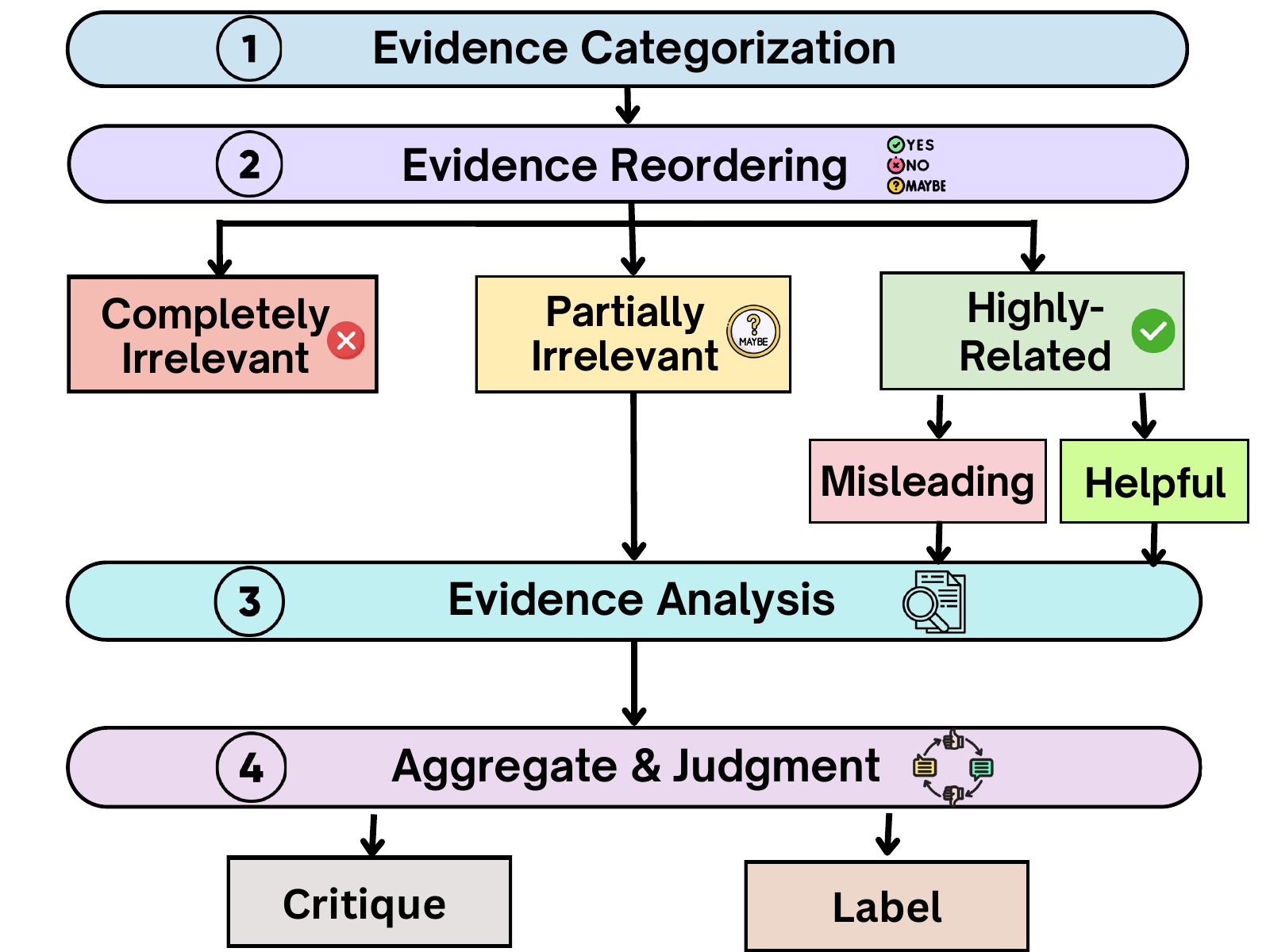}
  \caption{An overview of the framework for \titlename.} 
  \label{fig:framework_half}
\end{figure}

The propensity of Large Language Models (LLMs)~\cite{bubeck2023sparks, team2023gemini} to hallucinate
presents significant challenges to their reliability and widespread implementation in real-world applications~\cite{ji2023survey, zhang2023siren}. 
Current retrieval-based approaches for identifying hallucinations ~\cite{min2023factscore, chern2023factool} first gather pieces of evidence, which are then used to determine whether the content contains hallucination. Although these methods are somewhat effective, they encounter several major issues:
(i) \textbf{Lack of Detailed Explanations}: These techniques often lack detailed explanations for their detection results and do not assess the reliability of such explanations. This absence of interpretability diminishes the practical value of these detectors, especially in high-stakes situations. For instance, in medical settings, simply alerting a doctor to factual errors in generated patient information without providing evidence-backed explanations can erode trust in the system's outputs~\cite{xie2024doclens}.
(ii) \textbf{Deficiencies in Retrieval}: Many existing tools for detecting hallucinations depend heavily on LLMs~\cite{niu2024selfrefinement}, which can be misled by irrelevant data gathered by flawed retrieval systems~\cite{shi2023large, wang2023learning}, leading to incorrect assessments.
(iii) \textbf{Uniform Treatment of Evidence}: In real-world applications, substantiating claims often requires multiple evidence sources to ensure reliability and validity~\cite{kamoi2023wice, guo2022checking}. This highlights the importance of \textit{multiple-evidence hallucination detection} -- performing hallucination detection on a claim against multiple retrieved evidence. However, most hallucination detection framework treat all evidence uniformly, failing to differentiate between various types of sources. These challenges underscore the urgent need for a more reliable hallucination detection system, one that excels in handling multiple pieces of evidence and produces high-quality critiques. This improvement would significantly enhance the practical utility of hallucination detectors in real-world applications.

To address these challenges, we propose \titlename, an open-source, critique-based hallucination judge capable of handling complex, multiple-evidence scenarios (an overview of our framework is shown in Figure~\ref{fig:framework_half}). This system excels in generating high-quality critiques, categorizing evidence effectively, and integrating all relevant information to deliver precise hallucination detection. 
At the heart of \titlename are three key technical developments:
Firstly, we introduce the \texttt{ME-FEVER}, a pioneering dataset specifically designed for more reliable hallucination detection. Based on the foundational \texttt{FEVER} dataset~\cite{thorne2018fever}, \texttt{ME-FEVER} includes 3,901 instances that feature three types of evidence: \textit{completely irrelevant}, \textit{partially irrelevant}, and \textit{highly relevant}. 
The dataset is split into 2,663 training and 1,238 testing instances, providing a solid base for both training and evaluation purposes.
Secondly, we enhance \titlename with preference-based learning method~\cite{rafailov2023direct} 
to boost the system's ability to identify and prioritize relevant evidence, further enhancing the quality of the generated critiques.
Lastly, our evaluation strategy incorporates a comprehensive framework that assesses both the answer-level and critique-level performance of \titlename. This allows us to measure how effectively the system filters relevant evidence and produces quality critiques.

Our experiments demonstrate that \titlename outperforms all baseline models including GPT-4o under multiple-evidence hallucination detection setting and has close performance under single-evidence hallucination detection setting, as shown in \S~\ref{sec: results}. Additionally, the critiques generated by \titlename achieve evaluation scores close to those of GPT-4o and demonstrate the highest evidence-matching rate.

To summarize, our contributions are as follows:

\begin{itemize*}
\item We introduce \titlename, an open-source, critique-based hallucination detection model with 7 billion parameters capable of providing fine-grained critiques and filtering out unrelated information during hallucination detection.

\item We create \texttt{ME-FEVER}, a \textbf{m}ultiple-\textbf{e}vidence hallucination detection dataset based on \texttt{FEVER} that simulates real-world situations, offering a solid foundation for both training and evaluating systems on multi-evidence hallucination detection.

\item We establish a novel multiple-evidence hallucination detection workflow featuring evidence categorization, evidence reordering, evidence-by-evidence analysis, and information aggregation to enable the hallucination detection system to filter unrelated contexts and generate a reliable critique in the end.
\end{itemize*}

\section{Related Works}

\subsection{Critique Generation with LLMs}
The versatile generative capabilities of LLMs enable LLM-as-evaluators \cite{zheng2024judging, dubois2024alpacafarm} to generate natural language descriptions to evaluate the quality of model-generated content \cite{saunders2022self, sun2024metacritique}. While previous works on LLM evaluation have explored different methods for providing more reliable evaluations and critiques \cite{chiang2023vicuna, wang2023pandalm, li2023generative, sun2024metacritique, chern2024scaleeval}, none have focused on offering critiques with filter retrieval-augmentation for hallucination detection. We propose \titlename to address this gap.

\subsection{Retrieval-Augmented Hallucination Detection}
Earlier works on LLM-based retrieval-augmented hallucination detection systems \cite{min2023factscore, chern2023factool} focus on establishing a fine-grained framework for claim-level hallucination detection that leverages external knowledge or databases. More recent works on hallucination detection systems enable editing \cite{mishra2024fine}, improving efficiency \cite{tang2024minicheck}, and facilitating long-form fact-checking \cite{wei2024long}. Our work continues the effort to enhance retrieval-augmented hallucination detection systems by providing enhanced critique with filter retrieval augmentation.

\section{Preliminaries}
In this section, we define key terms and introduce our framework for retrieval-augmented hallucination detection.

\subsection{Key Terms}
\label{sec: key-terms}
We first define some key terms that are used throughout our paper.

\noindent\textbf{Prompt ($p$)} A query or instruction that users send to LLMs.

\noindent\textbf{Response ($r$)} A piece of text (usually in long-form) generated by the LLMs.

\noindent\textbf{Claim ($c$)} A verifiable statement extracted from the response.

\noindent\textbf{Label ($l$)} An answer that determines whether or not a claim $c$ is hallucinated, which can be \textit{True} (no hallucination), \textit{False} (with hallucination), or \textit{Neutral}.

\noindent\textbf{Evidence ($e$)} The available information or databases that could potentially help verify whether a claim $c$ is hallucinated or not.

\noindent\textbf{Critique ($cr$)} A natural language description for assessing whether a claim $c$ is hallucinated or not.

\subsection{Retrieval-Augmented Hallucination Detection Systems}
\label{sec: retrieval-augmented}
Previous retrieval-augmented hallucination detection frameworks \cite{min2023factscore, chern2023factool, wei2024long} typically consist of three primary components:
\begin{itemize*}
    \item \textbf{Claim Extraction}: Extracting fine-grained verifiable claims from a given response $r$.
    \item \textbf{Evidence Collection}: Utilizing retrieval tools or online search engines to retrieve external knowledge as evidence $e$.
    \item \textbf{Claim Verification}: Utilizing the evidence to verify whether a claim $c$ is hallucinated or not.
\end{itemize*}

\section{\texttt{ME-FEVER}: A Multiple-Evidence Hallucination Detection Dataset}
\subsection{Motivation}
To benchmark a hallucination detection system, one approach is to use standard natural language inference (NLI) datasets~\cite{thorne2018fever, nie2019adversarial} as test data. These datasets typically consist of a claim, a label, and a single piece of evidence for each sample. However, since each test sample includes only one piece of evidence, this is much simpler than the multiple-evidence hallucination detection scenarios that are often encountered in real-world applications. To address this limitation, we generate different types of evidence based on the \texttt{FEVER} dataset \citep{thorne2018fever}. By synthesizing them with the original \texttt{FEVER} data, we create \texttt{ME-FEVER}, a more challenging dataset used to train and benchmark hallucinations in models under multiple-evidence scenarios.

\subsection{Data Curation}
\label{Sec: ME-FEVER}
We prompt GPT-4-Turbo\footnote{gpt-4-turbo-2024-04-09} to generate multiple pieces of evidence based on the original \texttt{FEVER} evidence for each instance (detailed prompt is shown in Appendix \ref{sec:appendix_prompts}). The generated evidence is categorized into three predefined categories as follows:

\paragraph{Completely Irrelevant Evidence ($E^{o}$)} This type of evidence is entirely unrelated to the claims presented and should be disregarded during the hallucination detection process. In real-world scenarios, it may appear as a result of poorly formulated queries during retrieval, a lack of pertinent information in the knowledge base, or extensive evidence that includes unnecessary details. We randomly select two pieces of evidence from the \texttt{FEVER} dataset as completely irrelevant evidence. We manually compare them with the claims to ensure irrelevance.

\paragraph{Partial Irrelevant Evidence ($E^p$)} This type of evidence might appear related to the claim in subject matter or format, yet contribute minimally to the verification process. While this type of evidence may seem relevant, it often has limited impact on determining whether or not the claim is hallucinated. \titlename is designed to extract useful portions from this type of evidence and disregard the irrelevant parts. Therefore, we ask GPT-4-Turbo to create four separate paragraphs that match the subjects mentioned in the claims. These paragraphs neither contradict existing known facts nor specifically support or refute the claims. 
Since the paragraphs generated by GPT-4-Turbo are often short, we use GPT-3.5-Turbo\footnote{gpt-3.5-turbo-0125} to expand these paragraphs to approximately 150 words.

\paragraph{Highly Related Evidence ($E^r$)} This type of evidence is highly related to a claim. Note that each instance in \texttt{FEVER} contains only one piece of highly related evidence. To simulate more complex scenarios, we ask GPT-4-Turbo to generate three additional paragraphs of highly related evidence. These paragraphs are intended to be \textit{misleading}: they tend to mislead the hallucination detector into making an incorrect judgment about the claim, but not directly supporting or refuting whether the claim is hallucinated. These \textit{misleading} paragraphs are included in the dataset to help build a more challenging and robust dataset for hallucination detection. These misleading paragraphs must adhere to the following criteria:

\begin{enumerate*}
    \item They should not contradict the information present in the provided evidence.
    \item They should not violate the label of the claim, meaning they should neither support nor refute the claim.
    \item They should include confusing content to mislead the hallucination detector.
\end{enumerate*}

Despite explicit instructions, we note that there are still instances where the generated paragraphs conflict with the claims. Thus, we instruct GPT-4-Turbo to filter out such instances.

Overall, we synthesized a total of 3,901 instances for \texttt{ME-FEVER} based on the \texttt{FEVER} dataset. We generate additional pieces of evidence from the single evidence in \texttt{FEVER}. Of these, 2,663 instances form the training set, while the remaining 1,238 instances are used as the testing set. For each instance, there are two pieces of completely irrelevant evidence, four pieces of partially irrelevant evidence, and one to three pieces of highly related evidence.

\section{\titlename}




We introduce the hallucination detection framework for \titlename that mainly focuses on generating high-quality critiques and accurate prediction labels given a claim and a set of evidence more than one piece. Here, we outline our framework based on the key terms and concepts mentioned in \S\ref{sec: key-terms}, \S\ref{sec: retrieval-augmented}, and \S\ref{Sec: ME-FEVER}. 

\subsection{Problem Formulation}
We use evidence $E$ retrieved from external sources to effectively verify whether or not a claim $c$ is hallucinated. The main problem is to build a hallucination detector that:

\begin{enumerate*}
    \item Filters out irrelevant evidence in $E$.
    \item Provides a detailed critique of the claim $c$ based on the filtered evidence.
\end{enumerate*}

This process involves four distinct steps, detailed below.


\subsection{Framework}

\begin{figure*}[ht]
  \centering
  \includegraphics[width=\linewidth, height=8cm]{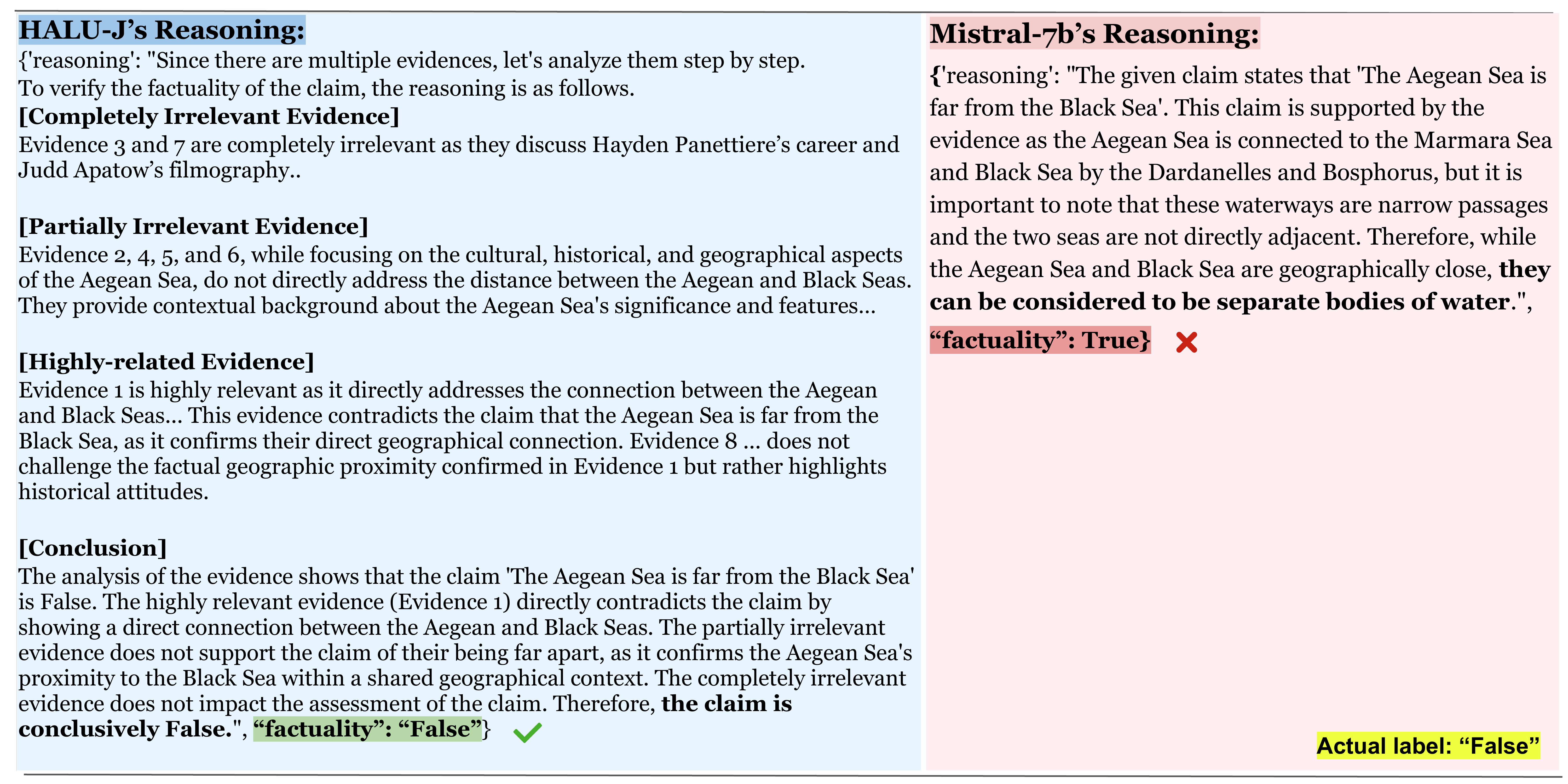}
  \caption{An example critique of \titlename vs. original Mistral-7b. Note that we use ``factuality'' as equivalent to ``no hallucination''} 
  \label{fig:example}
\end{figure*}

\subsubsection{Step-I: Evidence Categorization}
First, \titlename should systematically review all pieces of evidence and categorize each into one of the three predefined categories mentioned in \S\ref{Sec: ME-FEVER}.



We denote the $n$ pieces of evidence in $E$ as:
\begin{equation}
    E = \{e_1^{(t_1)}, e_2^{(t_2)}, \ldots, e_n^{(t_n)}\}
\end{equation}
where $ t_i \in \{o, p, r\} $ and $ i \in \{1, 2, 3, \ldots, n\}$.



\subsubsection{Step-II: Evidence Reordering}
\label{Sec:evidence_reordering}
\titlename should group the same types of evidence together to form different evidence groups, arranged in the following order: \textit{Completely Irrelevant Evidence, Partially Irrelevant Evidence}, and \textit{Highly Related Evidence}. By ordering these evidence, this ensures clarity and organization, which makes the detection process more manageable afterwards. Additionally, this standardized approach enhances accuracy and consistency by preventing models from overlooking any extracted evidence, thereby minimizing the likelihood of mistakes. This also allows models to think in a more systematic manner. We denote the evidence order as follows:

\[
E = \left\{
\begin{array}{l}
\underbrace{e_1^o, e_2^o, \ldots, e_{n_o}^o}_{E^o}, \\
\underbrace{e_1^p, e_2^p, \ldots, e_{n_p}^p}_{E^p}, \\
\underbrace{e_1^r, e_2^r, \ldots, e_{n_r}^r}_{E^r}
\end{array}
\right\}
\]

where $ n_o $ , $ n_p $ , and $ n_r $ denotes the number of evidence in $ E^o $ , $ E^p $ , and $ E^r $, respectively.

\subsubsection{Step-III: Evidence Analysis}
\label{Sec:analysis}
Our framework involves analyzing the relationships among the evidence and conducting a detailed analysis of how they each relate to the claim. \titlename should analyze and reason through each piece of evidence in a step-by-step manner. The analysis must meet the following requirements:

\begin{enumerate*}
    \item The analysis disregards completely irrelevant evidence.
    \item The analysis extracts the relevant parts in the partially irrelevant evidence, and discard the rest.
    \item The analysis clarifies how the helpful evidence (under highly-related evidence) support or refute the claim.
    \item If the evidence is identified as misleading, the analysis explains the relationship between the misleading evidence and the claims.
\end{enumerate*}
By clearly delineating the relevance of each piece of evidence, this method enhances the thoroughness of the hallucination detection process. Additionally, analyzing based on the categorized evidence streamlines the verification process, enabling \titlename to quickly dismiss irrelevant information and focus on the most relevant details. By placing the least relevant evidence at the beginning and the most relevant evidence toward the end of the generated analysis, this approach helps \titlename provide a clear, logical progression in their reasoning. We denote the progression in analysis as follows:

\[
A = \left\{ 
\begin{array}{l}
a_1^o, a_2^o, \ldots, a_{n_o}^o, \\
a_1^p, a_2^p, \ldots, a_{n_p}^p, \\
a_1^r, a_2^r, \ldots, a_{n_r}^r 
\end{array} 
\right\}
\]
where \( a_j^t \) corresponds to the analysis for each evidence \( e_j^t \).

\subsubsection{Step-IV: Aggregation and Critique Generation}
In this step, \titlename summarizes all the analysis and makes a conclusive determination on whether the given claim is \textit{true, false}, or \textit{neutral}. Then \titlename carefully checks whether a claim is hallucinated by evaluating whether it is supported by the most direct and relevant evidence available. This step is crucial as it ensures that all available pieces of evidence have been considered, allowing the detector to synthesize a coherent, well-founded, and reliable critique in the end. 

The generated critique includes detailed information such as the category of each piece of evidence, comprehensive analysis for each piece of evidence, and a concise conclusion judging the claim's hallucinations. This information aids in producing more accurate label prediction, which has never been used in past verification processes. 

To summarize, the final critique $cr$ and a corresponding label $l$ is generated using analysis $ A $. Overall, the process involves taking $ (c, E) $ as input and generates the corresponding $ (cr, l) $. An example critique is shown in Figure \ref{fig:example}.

\begin{table*}[ht]
  \centering
  \footnotesize
  \setlength{\tabcolsep}{12pt}
    \begin{tabular}{l|cccccc}
    \toprule
    \textbf{Model} & \multicolumn{1}{l}{ME-FEVER} & \multicolumn{1}{l}{FEVER} & \multicolumn{1}{l}{ANLI} & \multicolumn{1}{l}{WANLI} & \multicolumn{1}{l}{HaluEval} & \multicolumn{1}{l}{KBQA} \\
    \midrule
    \multicolumn{7}{l}{\textit{Closed-source Models}}\\
    \midrule
    GPT-3.5-Turbo & 0.81  & 0.87  & 0.47  & 0.47  & 0.59  & 0.69 \\
    GPT-4o & 0.83 & \textbf{0.88}  & \underline{0.74}  & \underline{0.60}  & \underline{0.81}  & \underline{0.84} \\
    \midrule
    \multicolumn{7}{l}{\textit{Open-source Models}}\\
    \midrule
    Mistral-7B-Instruct-v0.2 & 0.78  & 0.82  & 0.62  & 0.54  & 0.57  & 0.68 \\
    llama-2-13b-chat-hf & 0.13  & 0.37  & 0.27  & 0.29  & 0.24  & 0.19 \\
    Llama-3-8B-Instruct & 0.63  & 0.03  & 0.02  & 0.00  & 0.01  & 0.20 \\
    Qwen1.5-7B-Chat & 0.49  & 0.79  & 0.68  & 0.53  & 0.61  & 0.69 \\
    \titlename (w/o DPO) & \textbf{0.90}  & 0.90  & 0.69  & 0.54  & 0.65  & 0.76 \\
    \titlename & \underline{0.91}  & \underline{0.90}  & \textbf{0.70}  & \textbf{0.54}  & \textbf{0.65}  & \textbf{0.76} \\
    \bottomrule
    \end{tabular}%
  \caption{Label prediction accuracy of different models on various benchmarks. Results with \underline{underline} are the best among all models and results in \textbf{bold} are the second-best. \texttt{ME-FEVER} stands for the test set of multiple-evidence setting and \texttt{FEVER} stands for the test set under the single evidence setting.}
  \label{tab:performance-tasks}%
\end{table*}%

\subsection{Fine-tuning}

\subsubsection{Fine-tuning Data}
We use the following two types of data by taking a claim-evidence pair $ (c, E) $ as input and a critique-label pair $ (cr, l) $ as output for fine-tuning.

\paragraph{Multiple-Evidence Setting} 
We use the training set in \texttt{ME-FEVER} for fine-tuning in multiple-evidence setting. We keep the information on the type of each piece of evidence and clarifications of each misleading evidence during the evidence generation process detailed in \S\ref{Sec: ME-FEVER}. We then follow our framework through two stages: synthesization and reformatting (prompts used can be found in Appendix \ref{sec:appendix_prompts}).

For each instance, we prompt GPT-4-Turbo to synthesize a "golden" reasoning. This reasoning includes a detailed evidence-by-evidence analysis with the same requirements outlined in \S\ref{Sec:analysis}, classification results of the evidence, and a sound conclusion. The additional collected information and the exact label of the claim are also included in the synthesized prompt to ensure a high-quality critique and correct judgment.

Next, we prompt GPT-4-turbo to reformat the ``golden'' reasoning into a ``golden'' critique. The classification and ordering of evidence are inherently handled during reformatting. We further divide this set of instances into 1,952 instances for supervised fine-tuning and 709 instances for fine-tuning with DPO.


\paragraph{Single Evidence Setting} For the single evidence scenario, we generate 1,840 critiques from \texttt{FEVER} by directly prompt GPT-4-turbo with the exact label available in prompt.


\subsubsection{Fine-tuning Procedure} Following the above process, we obtain a set of prompt-response pairs under the multiple-evidence setting. The prompts include claims from FEVER and newly generated pieces of evidence, while the response contains the ``golden'' critiques and the corresponding labels for each claim. 

By integrating data from both single-evidence and multiple-evidence settings, we allow \titlename to seamlessly toggle between different hallucination detection scenarios, whether there is one piece of evidence or multiple. To reduce positional bias, we randomly shuffle the evidence in the multiple-evidence setting.

\paragraph{Fine-tuning with DPO}

To enhance the quality of critiques and improve the accuracy of the predicted label under the multiple-evidence setting, we further fine-tune \titlename (w/o DPO) with DPO \citep{rafailov2023direct} to obtain \titlename. 


\section{Experiments}
\label{sec: experiments}

\subsection{Experimental Setup}

\paragraph{Models} We use the models Mistral-7B-Instruct-v0.2 \cite{jiang2023mistral}, GPT-3.5-Turbo-0301, GPT-4o-2024-05-13, Llama-2-13b-Chat \citep{touvron2023llama}, Llama-3-8B-Instruct, and Qwen1.5-7B-Chat \citep{qwen1.5} for our baseline experiments. 

\paragraph{Fine-tuning Details}
For supervised fine-tuning, we obtain \titlename by fine-tuning Mistral-7B-Instruct-v0.2 \citep{jiang2023mistral} with DeepSpeed \citep{rasley2020deepspeed} library, Zero Redundancy Optimizer (ZeRO) \citep{rajbhandari2020zero,ren2021zerooffload} Stage 3, gradient-checkpointing \citep{chen2016training_gradckpt}, and FlashAttention \citep{dao2022flashattention,dao2023flashattention2} on 4 NVIDIA A100 GPUs. We use the bfloat16 (BF16) and tfloat32 (TF32) mix computation precision options to optimize efficiency. \titlename is trained for 20 epochs. We use AdamW \citep{loshchilov2017decoupled} as our optimizer with $\beta_1 = 0.9, \beta_2 = 0.95$ and weight decay of 0.1. We use a peak learning rate of 1e-5 with 10 warm-up steps, setting cosine learning rate decay to 0, a batch size of 16, and a maximum sequence length of 8,192. The loss is calculated only on the output end.

For DPO, we conducted inference 30 times using \titlename on the DPO training set with multiple-evidence data, with sampling parameters set to $\text{temperature}=1$ and $\text{top}\_p=0.9$. For each DPO training instance, we select an answer with correctly predicted label as the chosen answer and an answer with incorrectly predicted label as the rejected answer. In the DPO experiment, we set the learning rate to 1e-7 and trained for 3 epochs.

\subsection{Evaluation Setup}

\paragraph{Evaluating Accuracy on Hallucination Detection Tasks}

We use 1,000 instances each from ANLI \citep{nie2020adversarial}, WANLI \citep{liu2022wanli}, and HaluEval \citep{li2023halueval}, along with 233 instances from KBQA in FacTool \citep{chern2023factool} to evaluate the models' performance under the single evidence setting. On the other hand, we use the testing set (1,238 instances) from \texttt{ME-FEVER} to evaluate models' performance under the multiple-evidence scenario. We prompt the models to respond in a Python dictionary format. The dictionary should have two keys: "reasoning" and "factuality". They correspond to a critique and a label, respectively. Responses that do not follow the expected format and cannot be properly interpreted are considered incorrect. Response that has the same label as the reference label (either \textit{true, false}, or \textit{neutral}) is considered correct. Figure \ref{fig:example} showcases a comparison case between the critique generated by \titlename and Mistral-7b under \texttt{ME-FEVER}'s test set (complete example in Appendix \ref{sec:appendix_example}). \titlename correctly classifies all evidence for their corresponding categories and provides detailed reasoning that allows it to predict the label correctly.

\paragraph{Critique Evaluation by GPT-4-Turbo}
We utilize GPT-4-Turbo to rate the generated critiques under the multiple-evidence setting on a scale from 1 to 100 using a carefully designed prompt (see Appendix \ref{sec:appendix_prompts} for the full prompt). This prompt asks the model to first output a step-by-step reasoning process before providing a final score. Additionally, we conducted a study to evaluate the agreement rate between human annotators (authors of the paper) and GPT-4-Turbo. The human annotators asked to score the same set of critiques as GPT-4-Turbo, and we calculated the Pearson correlation to confirm the reliability of using GPT-4-Turbo's evaluations.

\paragraph{Evidence Matching Evaluation}
The ability to distinguish between relevant and irrelevant evidence is crucial for models to generate reliable critiques and thus produce more accurate predictions of labels in multiple-evidence settings. We thus measure the accuracy of a model correctly matching an evidence to its corresponding category in their responses using the \texttt{ME-FEVER} dataset. Models other than \titlename (w/o DPO) and \titlename are prompted with in-context demonstrations of the same critique format.

\paragraph{Evaluation with Response Formats}
We apply the same formatting requirements for the models other than \titlename (w/o DPO) and \titlename, which matches the format of our training data. We then test the models' label prediction accuracy on \texttt{ME-FEVER}. This experiment aims to evaluate the impact of having a response format on the models' performance in predicting labels.

\begin{table*}[ht]
\centering
\footnotesize
    \begin{tabular}{lccc}
        \toprule
        \textbf{Model} & \textbf{Accuracy} & \textbf{Critique Score} & \textbf{Evidence-Matching Rate}\\
        \midrule
        GPT-3.5-Turbo & 0.81 &  72.35 & 59.29\% \\ 
        GPT-4o & 0.83 & \underline{85.85} & 61.43\% \\ 
        Mistral-7B-Instruct-v0.2 & 0.78 & 61.30 & 51.22\% \\ 
        Llama-2-13b-Chat & 0.13 & 45.20 & 40.86\% \\ 
        Llama-3-8B-Instruct & 0.63& 76.15 & 47.57\% \\
        Qwen1.5-7B-Chat & 0.49 & 66.40 & 52.32\% \\ 
        \titlename (w/o DPO) & \textbf{0.90} & 82.60 & \textbf{66.89\%} \\ 
        \titlename & \underline{0.91} & \textbf{83.90} & \underline{68.11}\% \\ 
        \bottomrule
    \end{tabular}
    \caption{Models' evaluation results under multiple-evidence scenario. Results with \underline{underline} are the best among all models and results in \textbf{bold} are the second-best}
    \label{tab:model_scores}
\end{table*}

\subsection{Results and Discussions}
\label{sec: results}
\subsubsection{Results}


\paragraph{Hallucination Detection Accuracy} From Table \ref{tab:performance-tasks}, we see that our model, \titlename, significantly outperforms all other baseline models, including GPT-4o, under the multiple-evidence setting. Under single evidence setting, \titlename outperforms all models on \texttt{FEVER}'s test data and outperforms models other than GPT-4o in other test datasets. We notice that the accuracy of Llama-3-8B-Instruct is notably poor in certain datasets. This is because the outputs generated by this model in these cases cannot be interpreted into a Python dictionary correctly, which might be related to its ability of instruction following. We also notice that, though Mistral-7B-Instruct-v0.2 has only 7B parameters, the accuracy of its outputs perform quite well on many datasets.

\paragraph{Critique Evaluation} Table \ref{tab:model_scores} shows the results of the critique evaluation experiment. Our model, \titlename, has the second highest scores among all models. This experiment passes the entire response to the scorer without converting it to a Python dictionary, resulting in the Llama3-8b model demonstrating quite good quality in its generated critiques. This implies that despite the poor formatting performance of its responses, the quality of its critiques is significantly better.

Based on the annotated critique scores of 100 multiple-evidence \texttt{ME-FEVER} data, the Pearson correlation between humans and GPT-4-Turbo is 0.70, demonstrating a decent agreement between them.

\begin{figure}[ht]
  \centering
  \includegraphics[width=\linewidth]{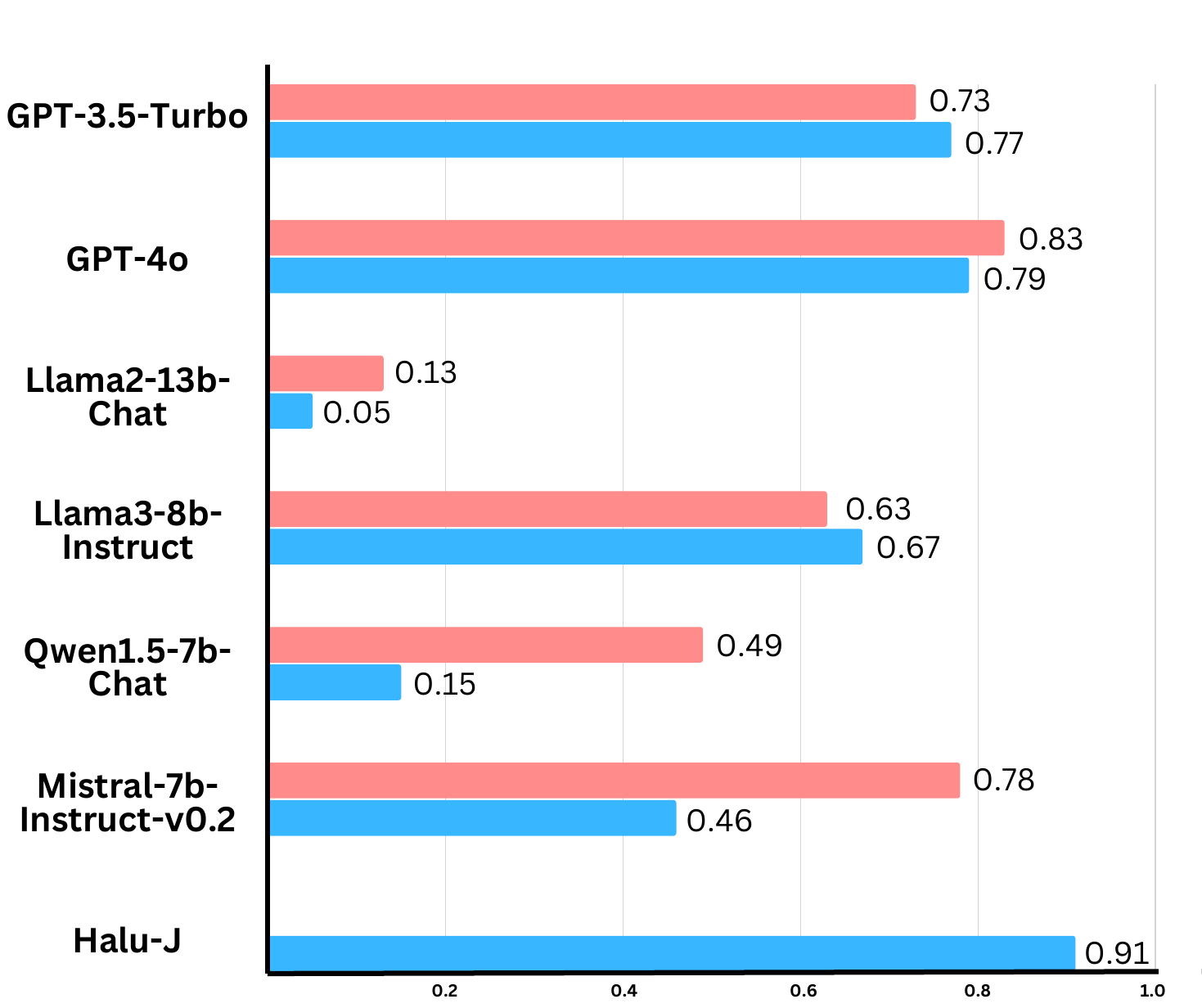}
  \caption{Label prediction accuracy comparison between models w/ and w/o formatting. Red bars are the performance results without formatting. Blue bars are the performance results with formatting.} 
  \label{fig:prompt_format}
\end{figure}

\paragraph{Evidence Matching} Table \ref{tab:model_scores} shows the results of the evidence matching experiment. \titlename outperforms all models on the evidence matching task, which means it has the best ability to classify and filter evidence. We notice that, in general, models with higher label prediction accuracy also tend to have higher evidence-matching rates. However, Mistral-7b  exhibits a notable discrepancy: it performs well in label prediction but falls short in evidence matching. This observation suggests that the ability to match evidence accurately is not necessarily essential for making accurate predictions.

\begin{table}[ht]
\centering
\scriptsize

    \begin{tabular}{@{\hskip 0.3em}c@{\hskip 0.7em}c@{\hskip 0.7em}c@{\hskip 0.5em}c@{\hskip 0.5em}c@{\hskip 0.5em}c@{\hskip 0.5em}}
        \toprule
        \textbf{Model} & ME-FEVER Acc & ANLI Acc & Crit. Score & Evi. Match.\\
        \midrule
        \titlename (w/o DPO) & 0.90 & 0.69  & 82.60 & 66.89\% \\ 
        \titlename & \textbf{0.91} & \textbf{0.70} & \textbf{83.90} & \textbf{68.11}\% \\ 
        \bottomrule
    \end{tabular}
    \caption{Comparison between \titlename (w/o DPO) and \titlename. Crit. Score stands for Critique Score. Evi. Match. stands for Evidence-Matching Rate. Results in \textbf{bold} are the best among all models.}
    \label{tab:DPO_compare}
\end{table}

\paragraph{Response Format} The results of the response format experiment is shown in Figure \ref{fig:prompt_format}. We observe that the response format do benefit some models such as GPT-3.5-Turbo and Llama-3-8b, but in most cases significantly reduces the accuracy of the models‘ label predictions. This suggests that expecting models to generate high-quality critiques solely with an in-context format template is impractical. The significant performance advantage of \titlename over Mistral-7b when using the response format demonstrates the effectiveness and necessity of our framework. By utilizing synthesized high-quality formatted data to train \titlename, we achieve superior results.


\subsubsection{Discussions}
\paragraph{Effectiveness of DPO Fine-tuning}

The results comparing \titlename (w/o DPO) and \titlename in Table \ref{tab:DPO_compare} demonstrate that DPO fine-tuning does help with performance improvement. We observe increases in accuracy on our \texttt{ME-FEVER} test set and under the ANLI dataset with single evidence setting. Additionally, there are improvements in critique scores and evidence-matching rates. This implies that DPO fine-tuning based on labels can enhance the overall quality of critiques, rather than merely increasing the accuracy of label predictions.


\section{Conclusion}
In this work, we develop \titlename, a hallucination detection judge with 7B parameters that verifies a claim based on given evidence. \titlename features its ability of generating high quality critique consisting of evidence categorization, detailed reasoning, and accurate label prediction under multiple-evidence real-world scenarios. We create the first multiple-evidence dataset \texttt{ME-FEVER} for hallucination detection, containing data from \texttt{FEVER} dataset and three kinds of synthesized evidence generated by GPT-4o. Experiments demonstrate that \titlename significantly outperforms open-source and closed-source baselines under both multiple-evidence and single-evidence hallucination detection tasks. Empirical results show the high quality of critique generated by \titlename. The resources in this work can facilitate future research on hallucination detection.

\clearpage
\section*{Limitations}
This work mainly focuses on commonsense reasoning and information-seeking hallucination tasks in LLM responses. Other types of hallucinations like numerical calculation errors are beyond our focus. \\
We carefully curated \texttt{ME-FEVER} as our training data to train our model's ability under multiple-evidence scenarios. However, there is much room for improvements in single-evidence scenarios.

\section*{Ethics Statement}
Our dataset \texttt{ME-FEVER} stems from \texttt{FEVER}, which is well-established and publicly available for use, containing no personal information. It is possible that \titlename can make mistakes. We urge users to double-check the hallucination detection results when using it in high-stakes scenarios. This work complies with the ACL Ethics Policy.


\bibliography{anthology,custom}
\bibliographystyle{acl_natbib}

\appendix

\newpage

\section{Prompts}
\label{sec:appendix_prompts}

\begin{table*}[t]
    \scriptsize
    \centering
\begin{tabular}{@{}p{\textwidth}@{}}
\toprule
\\
\begin{spverbatim}

I'll provide you with a claim and an associated evidence snippet. At the same time, you will be informed of the correctness of this claim. Your task is to output 4 other pieces of evidence related to the body of the claim, which has the following requirements:
1. The subject is the same as the subject in the claim
2. Do not contradict the information in the existing evidence
3. Don't support or oppose claim. This evidence should be a useless piece of information to judge the correctness of the claim
4. The length of the generated evidence should be close to the length of the provided evidence which is often  7-8 sentences
[claim]: {claim}
[evidence]: {evidence}
[correctness]: {label}

Here is your output format(a list of string with the length of 4):
["evidence1", "evidence2", "evidence3", "evidence4"]
Respond in the correct format directly.

\end{spverbatim}
\\
\bottomrule
\end{tabular}
    \caption{Prompt for generating partial irrelevant evidence.}
\end{table*}

\begin{table*}[t]
    \scriptsize
    \centering
\begin{tabular}{@{}p{\textwidth}@{}}
\toprule
\\
\begin{spverbatim}

I'll provide you with a claim and an associated evidence snippet. At the same time, you will be informed of the correctness of this claim. Your task is to output 3 other pieces of evidence related to the body of the claim, which has the following requirements:
1. Do not contradict the information in the existing evidence
2. The length of the generated evidence should be close to the length of the provided evidence, which is often  7-8 sentences
3. The evidence should not change the correctness of the claim, which is true. However, the evidence needs to contain confusing information to mislead the reader into believing that the claim is {opposite_label} (which is actually {label})
4. The misleading information needs to be deceptive enough.
5. Try to vary the pieces of evidence you generate and make them have different misleading points.

[claim]: {claim}
[evidence]: {evidence}
[correctness]: {label}

Here is your output format(a list of python dictionaries with the length of 3):
[{{
    "evidence": the evidence you generate,
    "explanation": What's the misleading information in the evidence(briefly)
}},...]

Respond in the correct format directly.

\end{spverbatim}
\\
\bottomrule
\end{tabular}
    \caption{Prompt for generating misleading evidence.}
\end{table*}

\begin{table*}[t]
    \scriptsize
    \centering
\begin{tabular}{@{}p{\textwidth}@{}}
\toprule
\\
\begin{spverbatim}

I'll provide you with a claim and associated evidence. At the same time, you will be informed of the correctness of this claim. Your task is to classify whether the evidence changes the correctness of the claim. If the correctness of the claim is neutral, it means either the supporting or the opposing evidence should be considered changing the correctness of the claim while the evidence neither supports or refutes the claim should be considered maintaining the correctness of the claim.

[claim]: {claim}
[label]: {label}
[evidence]: {evidence}

Here is your output format(a python dictionary):
[{{
    "explanation": the reason of your classification,
    "classification": True or False. True if the evidence maintains the correctness of the claim and False if the evidence changes the correctness of the claim
}}]

\end{spverbatim}
\\
\bottomrule
\end{tabular}
    \caption{Prompt for filtering misleading evidence.}
\end{table*}

\begin{table*}[t]
    \scriptsize
    \centering
\begin{tabular}{@{}p{\textwidth}@{}}
\toprule
\\
\begin{spverbatim}

I'm now training a large language model for claim verification. I expect it to generate high-quality critique given claim and evidence. Your task is to help me produce training datas.

Here is the given prompt I used for training:
[prompt used for training] 
You are given a claim. Your task is to identify whether there are any factual errors within the claim based on the given evidence.
The response should be a python dictionary with two keys - "reasoning", "factuality", which correspond to the reasoning and whether the given claim is factual or not (string - True, False or Neutral)
The following is the given claim
{claim}
The following is the provided evidences:
{formatted_evidences}

Now you'll get additional key information to help generate the 'golden' response expected.
{additional_info}
Most importantly, the factuality of the claim is {label}. Stick to this label when generating response.

Your task is to use the above additional information to output the 'golden' response as training datas, while making sure that you do not show that you got the extra information.

A 'golden' response requires the following requirements:
1. Be specific and complete in your response
2. Your response should start with an short assertion about the factuality of the claim like the factuality of the claim xxx is true / false / neutral
3. The response needs to go through each piece of evidence and analyze its relationship to the claim, which means you should analyze from the first evidence one by one
4. For evidence that is completely unrelated, you can briefly analyze it and point out that it is not related to the claim
5. For evidence that is basically unrelated to the content, you can briefly analyze and point out the reason like although it has the same subject as the claim, the content has nothing to do with the claim
6. For evidence with highly relevant content, give a reasonable analysis of whether the evidence support or oppose the claim, or neither support nor oppose it (take care to refer to the additional information I have provided you above, and do not show that you have this information. e.g. Be careful when using word like 'mislead'. The so-called misleading evidence is systhesized by me to mislead the chatbots. However, in real scenario, you should just regard it as a highly related but confusing evidence. You can not regard it as a wrong or misleading evidence)
7. When you go through the misleading evidence (which in real scenarios is highly related and confusing evidence) with explanation, you can consider include the information and discussion in the explanation in your analysis. You can also demonstrate the relevance between the claim and the evidence first.
8. At the end of the response, there needs to be a summary, which synthesizes the above analysis and derives the factuality of the claim

You should only respond in format as described below. DO NOT RETURN ANYTHING ELSE. START YOUR RESPONSE WITH '{{'.
{{
    "reasoning": "Why is the given claim factual non-factual or neutral? You must provide specific evidences to support your decision.",
    "factuality": "True" if the given claim is totally supported by the evidences, "False" if the given claim contradicts the evidences in some way, "Neutral" if evidence neither supports nor refutes the claim.
}}

\end{spverbatim}
\\
\bottomrule
\end{tabular}
    \caption{Prompt for synthesizing golden critique.}
\end{table*}

\begin{table*}[t]
    \scriptsize
    \centering
\begin{tabular}{@{}p{\textwidth}@{}}
\toprule
\\
\begin{spverbatim}

Given a claim, a set of evidence and a critique, your task is to reformat the critique.

CLAIM: {claim}
EVIDENCES: {evidence}
CRITIQUE: {critique}

The format of expected critique is as follows:

To verify the factuality of the claim, the reasoning is as follows.
[Completely Irrelevant Evidence]
A discussion and analysis of completely irrelevant evidence
[Partial Irrelevant Evidence]
A discussion and analysis of partial irrelevant evidence. The evidence must be analyzed case by case. You should point out the relevant and irrelevant information in the evidence respectively.
[Highly related Evidence]
A discussion and analysis of highly related evidence. The evidence must be analyzed case by case. You should dive into the details and discuss the relationship between the evidence and the claim.
[Conclusion]
Aggregate the analysis above and conclude whether the claim is true, false, or neutral

Your task is only reformat the critique. Don't change any reasoning or information in the original critique.
Output the reformatted critique directly.

\end{spverbatim}
\\
\bottomrule
\end{tabular}
    \caption{Prompt for reformatting golden critique.}
\end{table*}

\begin{table*}[t]
    \scriptsize
    \centering
\begin{tabular}{@{}p{\textwidth}@{}}
\toprule
\\
\begin{spverbatim}

I'm designing verifiers to judge whether a specific claim is correct based on evidence. There are three possible situations: Neutral / True / False. I am evaluating the quality of the verifier's responses. Your task is to score the response provided. The score range is 1-100. The ideal standard for responses are as follows:

The response should clearly point out whether the content of each piece of evidence is relevant to the claim.
The response should clearly identify the span in the claim that is particularly relevant to the relevant context in the evidence.
The response should clearly compare the related information in the claim and the evidence and provide a reasonable explanation. The explanation should be clear and reasonable.
The response should maintain faithfulness. It should not assert that the claim supports something not present in the original claim, nor should it suggest that the evidence supports something not found in the original evidence.
The response should ensure completeness, that is, all parts of the evidence that are highly relevant to the claim should be analyzed, and nothing should be omission.
The response should have a clear and reasonable logical reasoning process.

Here is the prompt for the verifier, which contains the claim and the evidence: 
{prompt}

Here is the response generated by the verifier: 
{response}

Try to be objective and start your response directly(low score for poor responses is encouraged)

You should only respond in format as described below. DO NOT RETURN ANYTHING ELSE. START YOUR RESPONSE WITH '{{'.
{{
    "reasoning": "a brief explanation of your score",
    "score": the score you provide
}}

\end{spverbatim}
\\
\bottomrule
\end{tabular}
    \caption{Prompt for scoring the critique.}
\end{table*}

\begin{table*}[t]
    \scriptsize
    \centering
\begin{tabular}{@{}p{\textwidth}@{}}
\toprule
\\
\begin{spverbatim}

I'll provide you with a set of evidence and a critique based on provided evidence.
Your task is to classify evidence into three categories which are [Completely Irrelevant Evidence], [Partial Irrelevant Evidence], and [Highly related Evidence].
You should strictly stick to the classification statements of the critique and don't change its meaning.
Evidence corresponding to the content in [Completely Irrelevant Evidence] should be matched to "Completely Irrelevant Evidence" category.
Evidence corresponding to the content in [Partial Irrelevant Evidence] should be matched to "Partial Irrelevant Evidence" category.
Evidence corresponding to the content in [Highly related Evidence] should be matched to "Highly related Evidence" category.
If there is evidence doen't correspond to any content in the critique, it should be matched to "Unmentioned" category.

Here is the evidence:

{evidence}

Here is the critique:

{critique}

Respond in a python dictionary with the following format:
{{
    "Completely Irrelevant Evidence":[evidence_number, evidence_number, ...],
    "Partial Irrelevant Evidence":[evidence_number, evidence_number, ...],
    "Highly related Evidence":[evidence_number, evidence_number, ...],
    "Unmentioned Evidence":[evidence_number, evidence_number, ...]
}}
Output in correct format directly.

\end{spverbatim}
\\
\bottomrule
\end{tabular}
    \caption{Prompt for extracting evidence-type dictionaries from critiques.}
\end{table*}

\clearpage

\section{Example}
\label{sec:appendix_example}

\begin{table*}[t]
    \scriptsize
    \centering
\begin{tabular}{@{}p{\textwidth}@{}}
\toprule
\\
\begin{spverbatim}

You are given claim. Your task is to identify whether there are any factual errors within the claim.
When you are judging the factuality of the given claim, you could reference the provided evidences if needed. The provided evidences may be helpful. Some evidences may contradict to each other. You must be careful when using the evidences to judge the factuality of the given claim.
The response should be a dictionary with two keys - "reasoning", "factuality", which correspond to the reasoning and whether the given claim is factual or not (string - True, False or Neutral)
The following is the given claim

[claim]: The Aegean Sea is far from the Black Sea.

The following is the provided evidences
[evidences]: 
1. Aegean Sea . In the north , it is connected to the Marmara Sea and Black Sea by the Dardanelles and Bosphorus .
/* golden evidence */ 

2. The Aegean Sea holds significant historical importance as it was the location for the early advancements of civilization in Europe. This body of water, located between the Greek and Anatolian peninsulas, played a critical role in the development and spread of culture, trade, and ideas throughout the region. Numerous archaeologically significant sites are situated along its shores, one of the most famous being the ancient city of Troy. The sea served as a vital maritime route that connected various civilizations, facilitating interactions and exchanges that greatly contributed to the cultural and economic growth during ancient times. Its strategic importance made it a central area for numerous historical events and narratives that shaped the course of European history.
/* partial irrelevant evidence */ 

3. Hayden Panettiere . She is known for her roles as cheerleader Claire Bennet on the NBC sci-fi series Heroes ( 2006 -- 10 ) , Juliette Barnes in the ABC/CMT musical-drama series Nashville ( 2012 -- present ) and Kairi in the video game series Kingdom Hearts . She began her acting career by playing Sarah Roberts on One Life to Live ( 1994 -- 97 ) , and Lizzie Spaulding on Guiding Light ( 1996 -- 2000 ) , before starring at age 10 as Sheryl Yoast in the Disney feature film Remember the Titans . She received two nominations for the Golden Globe Award for Best Supporting Actress -- Series , Miniseries or Television Film , for her work on Nashville in 2012 and 2013 .
/* completely irrelevant evidence */ 

4. The Aegean Sea, a significant and picturesque body of water, serves as an elongated embayment of the Mediterranean Sea, nestled between the Greek and Anatolian peninsulas. This sea is renowned for its historical and cultural significance, as it has been a crucial area for trade and communication among ancient civilizations. Within the Aegean Sea are the Aegean Islands, a group of islands that are not only geographically within the sea but also contribute to defining its southern boundary. Some of the most well-known of these islands include Crete and Rhodes, which are popular tourist destinations known for their stunning landscapes, rich history, and vibrant local cultures.
/* partial irrelevant evidence */ 

5. The Aegean Sea, a body of water located between the Greek mainland and Turkey, has a name with several possible origins. One theory suggests that the name comes from the ancient town of Aegae, which was situated near the sea. Another possibility is that the sea was named after Aegea, a mythical queen of the Amazons who is said to have perished in its waters. Additionally, the name could be derived from the Greek town of Aegae, further linking it to the region's history. Alternatively, the name might come from Aigaion, the "sea goat", which is another name for Briareus, a figure in Greek mythology. Each of these theories highlights the rich cultural and mythological significance of the Aegean Sea.
/* partial irrelevant evidence */ 

6. The Aegean Sea is a significant body of water situated between the Greek mainland and the coast of Turkey. Notably, the sea reaches a maximum depth of 3,543 meters within the Calypso Deep, which is situated in its northeastern sector. Overall, the sea spans a vast area, encompassing approximately 214,000 square kilometers. This regional expanse provides a critical habitat for diverse marine species and serves as an important route for maritime activities, including trade and transportation. The Aegean Sea is thus a geographically and economically significant area, playing a crucial role in the ecology and commerce of the surrounding regions.
/* partial irrelevant evidence */ 

7. Judd Apatow ( born December 6 , 1967 ) is an American film producer , writer , director , actor , and comedian . He is the founder of Apatow Productions , through which he produced and developed the television series Freaks and Geeks , Undeclared , Girls , Love , Crashing and directed the films The 40-Year-Old Virgin ( 2005 ) , Knocked Up ( 2007 ) , Funny People ( 2009 ) , This Is 40 ( 2012 ) , and Trainwreck ( 2015 ) . Apatow 's work has won numerous awards including a Primetime Emmy Award ( for The Ben Stiller Show ) , a Hollywood Comedy Award , and an AFI Award for Bridesmaids ( 2011 ) . His films have also been nominated for Grammy Awards , PGA Awards , Golden Globe Awards , and Academy Awards . Also known for producing hit films such as The Cable Guy ( 1996 ) , Anchorman : The Legend of Ron Burgundy ( 2004 ) , Talladega Nights : The Ballad of Ricky Bobby ( 2006 ) , Superbad ( 2007 ) , Forgetting Sarah Marshall ( 2008 ) , Step Brothers ( 2008 ) , Pineapple Express ( 2008 ) , Get Him to the Greek ( 2010 ) , Bridesmaids ( 2011 ) , The Five-Year Engagement ( 2012 ) , and Anchorman 2 : The Legend Continues ( 2013 ) .
/* completely irrelevant evidence */ 

8. Due to the historical significance of maritime trade in the region, the Aegean and Black Seas were often perceived as distant points, necessitating diverse trade routes to bridge these two important economic areas.
/* misleading evidence */ 

You should only respond in format as described below. DO NOT RETURN ANYTHING ELSE. START YOUR RESPONSE WITH '{'.
[response format]: 
{
    "reasoning": "Why is the given claim factual non-factual or neutral? You must provide specific evidences to support your decision.",
    "factuality": "True" if the given claim is totally supported by the evidences, "False" if the given claim contradicts the evidences in some way, "Neutral" if part of the given claim is supported by evidence, but there is also some information that cannot be determined as correct or incorrect.
}

\end{spverbatim}
\\
\bottomrule
\end{tabular}
    \caption{An prompt example of multiple-evidence scenario. We denote every evidence type with \\/* \{evidence\_type\} */ below each evidence. Evidence types don't appear in real prompts.}
    \label{tab:prompt_example}
\end{table*}

\begin{table*}[t]
    \scriptsize
    \centering
\begin{tabular}{@{}p{\textwidth}@{}}
\toprule
\\
\begin{spverbatim}

To verify the factuality of the claim, the reasoning is as follows.

[Completely Irrelevant Evidence]
Evidence 3 and 7 are completely irrelevant as they discuss Hayden Panettiere’s career and Judd Apatow’s filmography, which have no connection to the geographical relationship between the Aegean and Black Seas.

[Partially Irrelevant Evidence]
Evidence 2, 4, 5, and 6, while focusing on the cultural, historical, and geographical aspects of the Aegean Sea, do not directly address the distance between the Aegean and Black Seas. They provide contextual background about the Aegean Sea's significance and features but do not offer specific information regarding the spatial relationship with the Black Sea.

[Highly related Evidence]
Evidence 1 is highly relevant as it directly addresses the connection between the Aegean and Black Seas. It states that the Aegean Sea is connected to the Black Sea by the Dardanelles and Bosphorus, which clearly establishes a direct waterway between these two bodies of water. This evidence contradicts the claim that the Aegean Sea is far from the Black Sea, as it confirms their direct geographical connection.
Evidence 8, although discussing historical perceptions that might have considered the Aegean and Black Seas as distant, does not reflect the actual geographical relationship between these seas. It does not challenge the factual geographic proximity confirmed in Evidence 1 but rather highlights historical attitudes.

[Conclusion]
The analysis of the evidence shows that the claim 'The Aegean Sea is far from the Black Sea' is False. The highly relevant evidence (Evidence 1) directly contradicts the claim by showing a direct connection between the Aegean and Black Seas. The partially irrelevant evidence does not support the claim of their being far apart, as it confirms the Aegean Sea's proximity to the Black Sea within a shared geographical context. The completely irrelevant evidence does not impact the assessment of the claim. Therefore, the claim is conclusively False.

\end{spverbatim}
\\
\bottomrule
\end{tabular}
    \caption{The critique generated by \titlename based on the prompt in Table \ref{tab:prompt_example}}
\end{table*}

\begin{table*}[t]
    \scriptsize
    \centering
\begin{tabular}{@{}p{\textwidth}@{}}
\toprule
\\
\begin{spverbatim}

The given claim states that 'The Aegean Sea is far from the Black Sea'. This claim is supported by the evidence as the Aegean Sea is connected to the Marmara Sea and Black Sea by the Dardanelles and Bosphorus, but it is important to note that these waterways are narrow passages and the two seas are not directly adjacent. Therefore, while the Aegean Sea and Black Sea are geographically close, they can be considered to be separate bodies of water.

\end{spverbatim}
\\
\bottomrule
\end{tabular}
    \caption{The critique generated by Mistral-7B-Instruct-v2 based on the prompt in Table \ref{tab:prompt_example}}
\end{table*}

\end{document}